\documentclass[letterpaper, 10 pt, conference]{ieeeconf}  

\IEEEoverridecommandlockouts                              

\usepackage{cite}
\usepackage{amsmath,amssymb,amsfonts}
\usepackage{graphicx}
\usepackage{textcomp}
\usepackage{xcolor}
\usepackage{caption}
\usepackage{graphicx, subfig}
\usepackage{verbatim}
\usepackage[ruled,linesnumbered]{algorithm2e}
\usepackage{algorithmicx}
\usepackage{amsmath}
\usepackage{mathrsfs}
\usepackage{booktabs}
\usepackage{longtable}
\usepackage{marvosym}
\usepackage{balance}
\usepackage{color}
\usepackage{soul}
\usepackage{booktabs}
\usepackage{multirow}
\usepackage{caption}
\usepackage{array}
\usepackage{makecell}
\usepackage{mfirstuc}

\makeatletter
\let\NAT@parse\undefined
\makeatother  
\usepackage[colorlinks=false,linkcolor=black,anchorcolor=black,citecolor=black,CJKbookmarks=false]{hyperref}  

\newcommand{\texttildemid}{\ensuremath{\vcenter{\hbox{\texttildelow}}}} 

\title{\LARGE \bf
TactileAR: Active Tactile Pattern Reconstruction
}

\author{Bing Wu and Qian Liu$^{*}$
\thanks{This work was supported in part by the National Science Foundation of China(Grant No.62071083), and in part by the Dalian Science and Technology Innovation Foundation (No. 2022JJ12GX014).}
\thanks{All authors are with the Department of Computer Science and Technology, Dalian University of Technology, Dalian 116024, China. Emails: {wb1595946882@mail.dlut.edu.cn, qianliu@dlut.edu.cn}. }
\thanks{Corresponding author: Qian Liu}
}
\begin{document}

\maketitle
\thispagestyle{empty}
\pagestyle{empty}

\begin{abstract}
High-resolution (HR) contact surface information is essential for robotic grasping and precise manipulation tasks. However, it remains a challenge for current taxel-based sensors to obtain HR tactile information. In this paper, we focus on utilizing low-resolution (LR) tactile sensors to reconstruct the localized, dense, and HR representation of contact surfaces. In particular, we build a Gaussian triaxial tactile sensor degradation model and propose a tactile pattern reconstruction framework based on the Kalman filter. This framework enables the reconstruction of 2-D HR contact surface shapes using collected LR tactile sequences. In addition, we present an active exploration strategy to enhance the reconstruction efficiency. We evaluate the proposed method in real-world scenarios with comparison to existing prior-information-based approaches. Experimental results confirm the efficiency of the proposed approach and demonstrate satisfactory reconstructions of complex contact surface shapes. Code: \href{https://github.com/wmtlab/tactileAR}{https://github.com/wmtlab/tactileAR}

\end{abstract}

\section{Introduction \label{section:intro}}

Humans are naturally endowed with tactile hyperacuity \cite{human_tactile_sr}. For example, human touch can discriminate at a spatial resolution of \texttildemid 0.3 mm in braille reading, which is finer than the receptive field of individual touch receptors (\texttildemid 2 mm) in fingertips \cite{tactile_sr_2015}. However, for the tactile sensing in robotics, it is still a challenging task for low-resolution (LR) taxel-based tactile sensors to meet the demand of high-resolution (HR) tactile data. 

In this context, several approaches have been developed to obtain HR and high-quality tactile data. One common method is to enhance the physical spatial resolution of tactile sensors. For example, vision-based sensors \cite{gelsight_2017_sensor, gelslim_2018_IROS, tactip_2018_SoftRobot} employ cameras to capture the deformation of sensor surface, which can detect the tactile information at the fingerprint level and provide HR tactile data \cite{gelsight_2017_sensor, gelslim_2018_IROS}. However, this type of sensors tend to be bulky due to the space required by the deployment of imaging equipment. As a result, research topics related to the miniaturization \cite{vision-sensor-2022, tactip-review-2020}  and flexibility\cite{soft_vit_2022_RoboSoft} of vision-based sensors attract increasingly more attentions nowadays. 

Another type of tactile sensors, known as taxel-based sensors, consists of an array of small sensing elements called taxels. Each taxel can measure one- or three-axis deformation information within a specific contact area. These sensors are available in various sizes and easier to integrate with existing robotic systems. However, the resolution of taxel-based sensors is generally much lower than that of vision-based counterparts. Efforts\cite{mems-sensor-2008, mems-sensor-2015} have been made to increase the taxel density. Nonetheless, various issues emerge with the increase of the taxel density such as more wire connections, lower response frequencies, and amplified cross-talk between taxels\cite{tactile_sensor_review_2015_RAS, tactile_sr_2022_ral}.

\begin{figure}[t]
	\centering
	\includegraphics[width=0.45\textwidth]{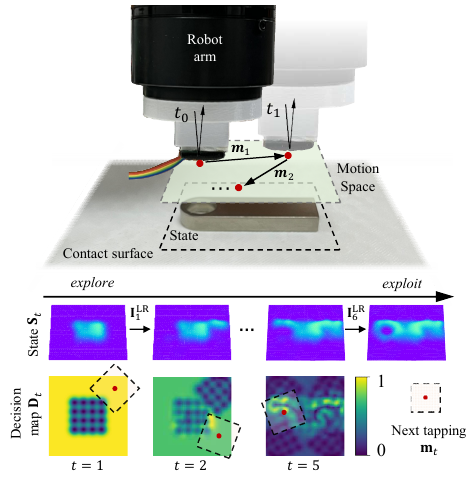}
	\caption{An illustration of proposed active tactile pattern reconstruction process. The tactile sensor taps the contact surface vertically downward and reaches a given height to collect a LR tactile pattern $\mathbf{I}^{\mathrm{LR}}_t$. The collected LR data is used to update the $\mathbf{S}_t$ , which is defined as the 2D shape of contact surface.  In the first several taps, the sensor tends to prioritize unexplored areas (e.g. $t=1, 2$). As the number of taps $t$ increases, the sensor gradually focuses on the contour information of the reconstructed surface (e.g. $t=5$).
 }
	\label{tap_system}
        \vspace{-0.6 cm} 
\end{figure}

An alternative solution is to improve the resolution of taxel-based sensors with fine designed algorithms. For example, \cite{tactile_sr_2015, tactile_sr_2015_iros, tactile_sr_2022_sr, tactile_sr_2022_ral} were able to enhance the localization accuracy of the contact point when single or multiple stimuli were applied. Given the similarities between image and tactile data, \cite{wubing_tactile_sr_iros_2022, wubing_tactile_sr_toh_2023} proposed the tactile pattern super resolution (SR),  which leveraged the prior information and LR tactile sequence to predict the HR tactile data. However, this method heavily relied on high-quality training datasets and unfortunately cannot achieve satisfactory performance when contact surfaces were complex or had a considerable difference from these in the training dataset. 

In this paper, we propose an active tactile pattern reconstruction framework, named TactileAR, which mimics the human ability to predict the shape of an unknown object through continuous touch. The TactileAR collects LR tactile data by continuously tapping the contact surface, then reconstructs the HR shape of the contact surface without using any prior information of the shape. This way, there is no need for the TactileAR to train a neural network model in advance, which also avoids the generalization problem of learning-based SR models \cite{wubing_tactile_sr_iros_2022, wubing_tactile_sr_toh_2023}. In particular, the proposed tactile pattern reconstruction framework is based on the Kalman filter. We first introduce  an explicit Gaussian tactile sensor degradation model, which depicts the degradation of HR tactile data into LR X-, Y-, and Z-axis observation data which can be collected by the raw tactile sensor, then reconstructs the 2D HR shape by using LR tactile sequences and the position information with the Kalman filter.

Furthermore, we propose an active tapping strategy inspired by the human tactile exploration process to improve the reconstruction efficiency. We discover that when facing an unfamiliar 2-D shape, humans tend to initially determine the position of the shape, then use their fingers for contour tracing with detailed and continuous touch. This process promotes the decrease of uncertainty and also presents the importance of the contour information. Therefore, we develop a task-driven active tactile exploration policy, which a heuristic strategy considering both the contour and the uncertainty of contact surfaces. From Fig. \ref{tap_system}, we can observe that the sensor tends to explore unvisited areas in the first few taps, and gradually focuses on the contour information of the reconstructed surface as the number of taps increases. 

\section{Relate Work}

Using a limited and sparse tactile sequence to obtain spatial features of a contact object is an important task in robot manipulation. These spatial features are useful in object recognition, pose estimation and dexterity manipulation\cite{tactile_pattern_review_2017, tactile_perception_review_2020_TRO}. In this context, two key components are involved, including the effective data acquisition policy, i.e. the tactile exploration, and how to reconstruct the object's spatial information from acquired tactile data, i.e. the tactile reconstruction.

The tactile exploration generally represents the data acquisition policy of tactile sensors, consisting of passive and active exploration policies \cite{tactile_active_2013_rss}. Passive exploration policies generally generate sampling trajectories by either fixed or randomized methods. For example, a fixed trajectory is used in \cite{tactile_mapping_2011, wubing_tactile_sr_iros_2022} to explore the contact surface, which is  time-consuming and inefficient. Conventional active exploration algorithms focus on reducing the uncertainty\cite{active_pure_2016_iros, active_pure_2013_ICRA } and generally collect information about areas that have not been explored yet, which is revealed by heuristic exploration approaches \cite{active_heur_tactile_2022_tro} with potential efficiency improvements in tactile exploration. \cite{active_heur_tactile_2011_tro, active_heur_tactile_2017_ral ,active_heur_tactile_2022_tro} developed task-oriented heuristic algorithms with higher efficiency compared with the conventional approach. Reinforcement learning methods, employing a trained agent to generate the exploration policy, have received increasingly high attentions \cite{active_rl_tactile_2020_ral, active_rl_tactile_2022_iros}, but unfortunately also face sim2real challenges \cite{tactile_sim_to_real_review}.

The tactile reconstruction indicates the process of using sensor-collected tactile data to predict or reconstruct the shape of contact surfaces. As for vision-based tactile sensor, \cite{vision_tactile_map_2019_ICRA, DenseTact_2022_ICRA} reconstructs the dense depth information of the contact surface from a single HR RGB image with image processing techniques.  \cite{vision_tactile_reconstruction_2018_IROS} proposed a deep learning model to reconstruct the full 3D shape of object via the visual information, the tactile information and learned shape priors. Different from HR vision-based sensors, the LR nature of taxel-based sensors makes it necessary to execute multiple times or fuse information from other modality (e.g., visuals) to reconstruct the object's spatial information. For example,  \cite{visual_taxle_reconstruction_2013_IROS, visual_taxle_reconstruction_2014_IJRR} use visual observations to obtain a coarse 3D shape of the object based on some assumptions, then use tactile sensors to refine the 3D shape.  \cite{tactile_reconstruction_2011_TRO_short} utilizes the Kalman filter to build a probabilistic model of the contact point cloud, and reconstructs a sparse 3D point cloud of the object. \cite{GPIS_2017_IROS, GPIS_2019_ICRA} use the Gaussian process implicit surface model to represent the shape of object and update the model by sliding over the surface with touch probes. 

In summary, most of existing work reconstructs the global 3D spatial information of the contact object, or directly infers the tactile properties of the object (e.g., object categories, and texture types) from the raw data. In this paper, we focus on the local spatial information of the contact object, and only use the LR tactile data to reconstruct the HR (dense) shape information of the corresponding region. We believe that this research can be considered as a valuable reference to the field of robot grasping and manipulation.

\section{Tactile Pattern Reconstruction \label{section:tactileAR}}
In this section, we present the proposed reconstruction framework which is able to obtain the HR shape of the contact surface using LR tactile sequences. First, we establish the tactile sensor observation formula, which is a Gaussian degradation model. Then, based on the Kalman filter and the observation formula, we derive the update formula for the state (i.e., the 2D shape of the contact surface).

\begin{figure}[t]
	\centering
	\includegraphics[width=0.4\textwidth]{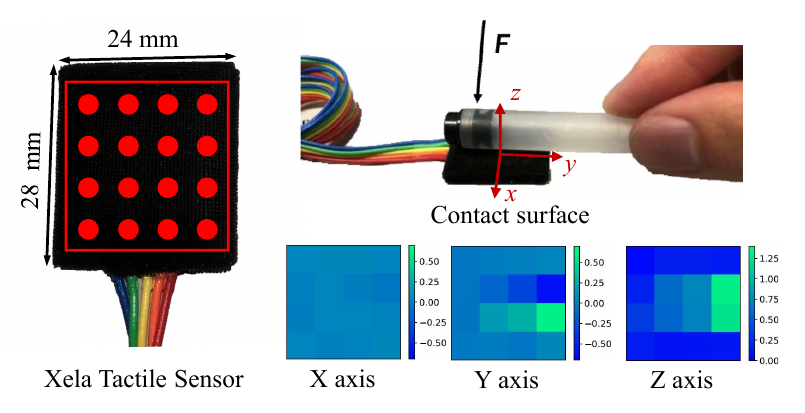}
	\caption{An example of 3-axis LR tactile signals collected by the Xela tactile sensor. The sensor has 4 $\times$ 4 taxels with \texttildemid 20$\times$20 mm$^2$ sensing area (adapted from \cite{wubing_tactile_sr_iros_2022}).
 }
	\label{fig:xela_tactile_sensor}
            \vspace{-0.6 cm}  
\end{figure}

\subsection{Problem Statement}
The tactile sensor needs to sample the contact surface continuously, with the number of tappings denoted as $t$, resulting in a set of observed sequences $[\mathbf{I}^{\mathrm{LR}}_0, \mathbf{I}^{\mathrm{LR}}_1, \cdots, \mathbf{I}^{\mathrm{LR}}_t]$ and position information $[\mathbf{m}_0, \mathbf{m}_1, \cdots, \mathbf{m}_t]$, as illustrated in Fig. \ref{tap_system}. Our objective is to reconstruct HR shape for the sampled region with these information.

The shape of tactile sensor that we use in this paper is square as shown in Fig.\ref{fig:xela_tactile_sensor}
, but the proposed approach is applicable to planar sensors of general shapes. Suppose the sensor data $\mathbf{I}^{\mathrm{LR}}_t$ is composed of $N\times N$ elements, with the sensor size of $l^{\mathrm{sensor}}\times l^{\mathrm{sensor}}$. Consequently, the resolution is $d^{\mathrm{LR}}=\frac{N}{l^{\mathrm{sensor}}}$. We postulate an HR ideal sensor data $\mathbf{I}^{\mathrm{HR}}_t$, with the same size as the actual sensor, consisting of $M\times M(M>N)$ elements, then the resolution is $d^{\mathrm{HR}}=\frac{M}{l^{\mathrm{sensor}}}$. We define the shape of the reconstruction area as the system state $\mathbf{S}_t$, with the size $l^{\mathrm{state}}=\alpha l^{\mathrm{sensor}}$ and the resolution identical to the HR ideal sensor data, comprising $\alpha M \times \alpha M$ elements. For descriptive purposes, we flatten  $\mathbf{S}_t \in \mathbb{R}^{(\alpha \cdot M)^2\times 1}$ ,  $\mathbf{I}^{\mathrm{HR}}_t \in \mathbb{R}^{M^2\times 1}$ ,  $\mathbf{I}^{\mathrm{LR}}_t \in \mathbb{R}^{N^2\times 1}$ into 1D representations.  

\subsection{Tactile Sensor Degradation Model}

Assuming the contact surface is fixed and not deformable throughout the tapping process. The state $\mathbf{S}_{t} $, i.e. the shape of the contact surface, does not change with the sensor movement. Therefore, the state equation of the system is

\begin{equation}
	\mathbf{S}_{t+1} = \mathbf{I} \cdot \mathbf{S}_{t},
    \label{eq:state_func}
\end{equation}
where $\mathbf{I}$ is the identity matrix. Eq.\ref{eq:state_func} means that we can seamlessly utilize the posterior of state at $t$-th as the prior of state at $t+1$-th tapping.

\begin{figure}[t]
	\centering
	\includegraphics[width=0.4\textwidth]{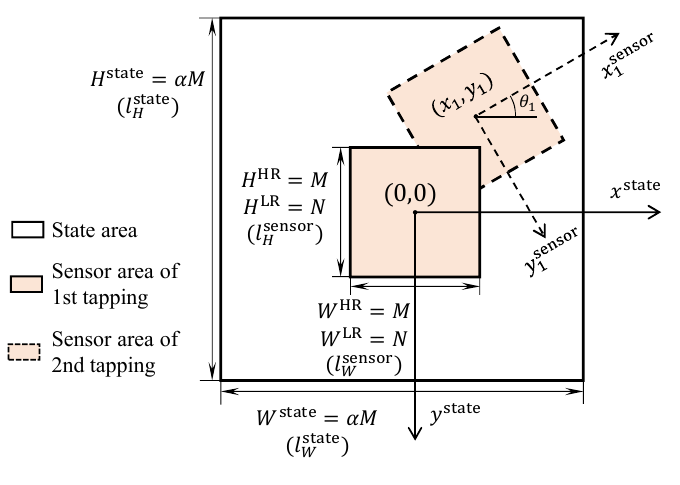}
	\caption{Problem description. Suppose the LR tactile sensor has $N\times N$ taxels with $l^{\mathrm{sensor}}\times l^{\mathrm{sensor}}$ sensing area. Assume that a HR sensor has the same size as the LR sensor, but with $M\times M$ taxels. The data collect by the LR sensor can be considered as the degradation of the HR sensor. The reconstructed area is $\alpha$ times larger than the sensor's sensing area, but has the same resolution as the HR sensor. The sensor center of the first tapping is taken as the origin of the reconstruction system.}
	\label{tactile_space}
        \vspace{-1em} 
\end{figure}

\noindent \textbf{Clip Matrix}: The Z-axis HR data $\mathbf{I}^{\mathrm{HR}}_{z, t}$ of the tactile sensor can be regarded as obtained from the state $\mathbf{S}_t$ by translation, rotation and clipping, as illustrated in Fig.\ref{tactile_space}. We define this process as
\begin{equation}
    \begin{split}
	\mathbf{I}^{\mathrm{HR}}_{z, t} &= \mathcal{F}(\mathbf{S}_t, \mathbf{m}_t) \\ 
             &= \mathbf{C}_t \cdot \mathbf{S}_t 
    \end{split},
	\label{eq:clip_matrix}
\end{equation}
where $\mathbf{m}_t$ is the motion parameter at the $t$-th tapping.It describes the position of the sensor relative to the initial pose. In this paper, we consider the 2-D motion of the sensor, i.e. $\mathbf{m}_t=(x_t, y_t, \theta_t)$, where $x_t$, $y_t$ and $\theta_t$ denote the translational offsets along the X-axis, Y-axis, and rotational offset around the Z-axis, respectively. The clip matrix, denote as $\mathbf{C}_t$ , elucidates how the state $\mathbf{S}_t$ are transformed to the HR tactile data $\mathbf{I}^{\mathrm{HR}}_t$, uniquely determined by the motion parameter $\mathbf{m}_t$. 

Expanding Eq.\ref{eq:clip_matrix}, we get

$$
    \begin{bmatrix}
        I^{\mathrm{HR}}_{z,t, 0} \\
        \vdots \\
        I^{\mathrm{HR}}_{z, t, i} \\
        \vdots\\
        I^{\mathrm{HR}}_{z, t, M^2}
    \end{bmatrix}
    = 
    \begin{bmatrix}
        C^{0,0}_t  & \cdots                   &  C^{0,(\alpha M)^2}_t \\
                                & \vdots                  &                         \\
           \cdots               &  C^{i,j}_t  &  \cdots \\
                                &   \vdots                &   \\
        C^{M^2,0}_t & \cdots                  &  C^{M^2,(\alpha M)^2}_t \\
    \end{bmatrix} 
    \begin{bmatrix}
        S_{t, 0} \\
        \vdots \\
        S_{t, j} \\
        \vdots\\
        S_{t, (\alpha M)^2}
    \end{bmatrix},
$$
where $C^{i,j}_t$ is an indicator function. $C^{i,j}_t=1$ indicates that ${S}_{t,j}$ , the $j$-element in $\mathbf{S}_t$, is in the same location as $I^{\mathrm{HR}}_{z,t, i}$, the $i$-element in  $\mathbf{I}^{\mathrm{HR}}_{z,t}$, otherwise, $C^{i,j}_t=0$. We define three symbols to represent location information $\mathbf{v}_j, \mathbf{v}^{t}_i, \mathbf{u}^{t}_i \in \mathbb{R}^2$. Specifically, $\mathbf{v}_j$ and $\mathbf{v}^{t}_i$ describe the location of state ${S}_{t,j}$ and HR $I^{HR}_{z,t, i}$  in their respective coordinate system respectively, while  $\mathbf{u}^{t}_i$  represents the location of the HR data $I^{HR}_{z,t, i}$ in the state coordinate system. We need to transform the location of HR data into the state coordinate system for comparison.

\begin{figure}[t]
	\centering
	\includegraphics[width=0.4\textwidth]{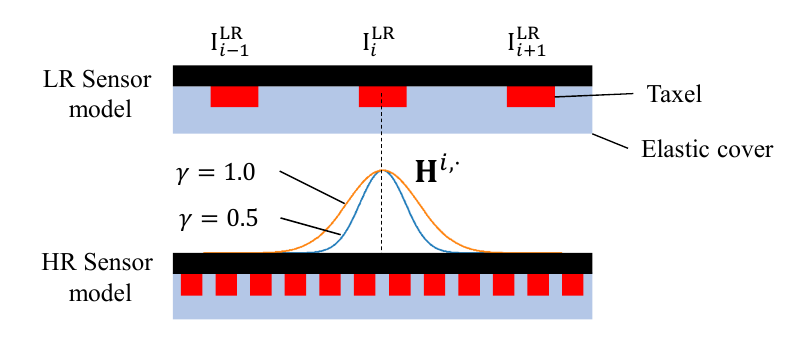}
	\caption{An illustration of the degradation process from the HR tactile data $\mathbf{I}^{\mathrm{HR}}_t$ to LR data $\mathbf{I}^{\mathrm{LR}}_t$. The HR sensor model is an idealized representation with smaller taxel size and smaller distances between adjacent taxels than the actual LR sensor (adapted from \cite{wubing_tactile_sr_toh_2023}).
        }
	\label{fig:degradation_model}
        \vspace{-1.0 em} 
\end{figure}

\begin{equation}
    \begin{split}
	 \mathbf{u}^{t}_i = \mathbf{R}^{t} \cdot \mathbf{v}^{t}_i +  
                \begin{bmatrix}
                    x_t \\
                    y_t
                \end{bmatrix}
    \end{split}, 
	\label{eq:rotate_eq}
\end{equation}
where $\mathbf{R}^{t}$ is the rotation matrix

\begin{equation}
\mathbf{R}^{t}= 
\begin{bmatrix}
    \cos \theta_t & -\sin \theta_t \\
    \sin \theta_t & \cos \theta_t
\end{bmatrix}.
\end{equation}

Eq.\ref{eq:rotate_eq} transforms the HR data $\mathbf{I}^{HR}_t$ into the same coordinate system as $\mathbf{S}_t$, enabling the comparison of location between these two parameters. For the clipping operation, we assign the tactile data of $\mathbf{S}_t$ to $\mathbf{I}^{\mathrm{HR}}_t$ of the same location. This way, we have

\begin{equation}
    C^{i,j}_t = 
    \exp\left\{ - \frac{||\mathbf{v}_j - \mathbf{u}^t_i||^2}{\beta_C} \right\} 
    \frac{1}{
    \Sigma^{(\alpha M)^2}_{j=0} C^{i,j}_t
    }.
\label{eq:sim_indicator_func}
\end{equation}

Eq.\ref{eq:sim_indicator_func} can approximate the indicator function when  $\beta_C$ is small enough.  In this paper, we assume $\beta_C = 1e-3$.

\noindent \textbf{Degradation Matrix}: The LR data $\mathbf{I}^{\mathrm{LR}}$ collected by the sensor can be perceived as HR data $\mathbf{I}^{\mathrm{HR}}$ derived through a degradation process with Gaussian noise.

\begin{equation}
    \begin{split}
	\mathbf{I}^{\mathrm{LR}}_t = \mathcal{D}(\mathbf{I}^{\mathrm{HR}}_t, \gamma) + \boldsymbol{\epsilon}_{t}  
    \end{split}, 
    \label{eq:degradation}
\end{equation}
where $\boldsymbol{\epsilon}_{t}  \sim \mathcal{N}(0, \mathbf{Q}_t)$, and $\mathbf{Q}_t$ is a diagonal matrix that describes the sensor noise. $\mathcal{D}$ represents the degradation process, which describes how HR tactile data is transformed into LR data, and also serves as the measurement model of the sensor. In this paper, we assume that the raw X-, Y- and Z-axis LR tactile data are derived from the Gaussian degradation of the corresponding HR data. The Gaussian degradation is an explicit linear degradation description, extensively employed in the computer vision research area \cite{cv-gauss-degradation-cvpr-2018}, where the values of the LR taxel are obtained from the HR taxel by Gaussian weighting. Therefore, we rewrite Eq.\ref{eq:degradation} with the Gaussian degradation matrix as

\begin{equation}
    \begin{split}
        \mathbf{I}^{\mathrm{LR}}_{x, t} &= \mathbf{H}(\gamma_x) \cdot \mathbf{I}^{\mathrm{HR}}_{x, t} +\boldsymbol{\epsilon}_{x,t} \\
	\mathbf{I}^{\mathrm{LR}}_{y, t} &= \mathbf{H}(\gamma_y) \cdot \mathbf{I}^{\mathrm{HR}}_{y, t} +\boldsymbol{\epsilon}_{y,t} \\
	\mathbf{I}^{\mathrm{LR}}_{z, t} &= \mathbf{H}(\gamma_z) \cdot \mathbf{I}^{\mathrm{HR}}_{z, t} +\boldsymbol{\epsilon}_{z,t} \\   
    \end{split}
    \label{eq:degradation_matrix},
\end{equation}
where $\mathbf{H}(\gamma)$ is the degradation matrix, which is established once the sensor is determined. According to Eq.\ref{eq:degradation_matrix}, $\mathbf{H}(\gamma) \in \mathbb{R}^{N^2 \times M^2}$, where $H^{i,j}(\gamma)$ signifies the effect of the $j$-th element in $\mathbf{I}^{\mathrm{HR}}_t$  to the $i$-th element in $\mathbf{I}^{\mathrm{LR}}_t$.

\begin{equation}
    \begin{split}
    H^{i,j}(\gamma)   = 
    \exp\left\{ - \frac{||\mathbf{v}^{HR}_j - \mathbf{v}^{LR}_i||^2}{\gamma} \right\} 
    \frac{1}{
    \underset{j}{\max} \ H^{i,j}(\gamma)
    }
    \end{split},
    \label{eq:degradation_gamma}
\end{equation}
where $\gamma$ is the degradation parameter, which represents each taxel's perceptual region during the degradation process. In this paper, we assume that each axis possesses a single parameter describing the degradation process, denoted as  $\gamma_x, \gamma_y, \gamma_z$. For more elaborate degradation processes, individual degradation parameter for each taxel is also feasible. We have observed that the reconstruction performance of the TactileAR is not highly sensitive to the degradation parameters. Through practical experimentation, $\gamma_x = \gamma_y = 1.0 , \gamma_z=2.0$ are adopted in this paper.

For the tactile pattern, the X-, Y- and Z-axis tactile data exhibit a distinct physical relationship. When the tactile sensor is in vertical contact with the contact surface, i.e., no additional tangential force is applied. The X- and Y-axis data can be regarded as the derivative of the Z-axis data in the X and Y direction, respectively. Hence, we have 

\begin{equation}
    \begin{split}
        \mathbf{I}^{\mathrm{HR}}_{x,t} = \nabla_x \ \mathbf{I}^{\mathrm{HR}}_{z, t} = \mathbf{G}_x \ \mathbf{I}^{\mathrm{HR}}_{z, t} \\
        \mathbf{I}^{\mathrm{HR}}_{y,t} = \nabla_y \ \mathbf{I}^{\mathrm{HR}}_{z, t} = \mathbf{G}_y \ \mathbf{I}^{\mathrm{HR}}_{z, t} \\
    \end{split},
    \label{eq:xy_gradient}
\end{equation}
where $\nabla_x$, $\nabla_y$ indicate the gradient operators. Since $\mathbf{I}^{\mathrm{HR}}_{z, t}$  is reshaped from a 2-D image to a 1-D vector in this process, we express the Sobel operator in the matrix form as $\mathbf{G}_x, \mathbf{G}_y \in \mathbb{R}^{M^2 \times M^2}$ to compute the gradients.

\subsection{State Updates}
We calculate the posterior probability of the state $\mathbf{S}_t$ based on the state equation (Eq.\ref{eq:state_func}), observation equations (Eq.\ref{eq:clip_matrix}, \ref{eq:degradation_matrix}, \ref{eq:xy_gradient}), and the prior probability. Assuming the initial state, $\mathbf{S}_0$, conforms to the Gaussian distribution, we have

\begin{equation}
    \begin{split}
	\mathbf{S}_0 \sim \mathcal{N}(\mathbf{\mu}_0, \mathbf{\Sigma}_0) \\ 
    \end{split}.
\end{equation}

\begin{figure}[t]
	\centering
	\includegraphics[width=0.48\textwidth]{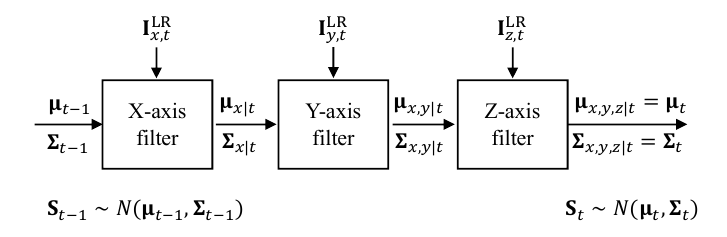}
	\caption{An illustration of the Sequential filtering process of the Kalman filter. The triaxial data gathered by the sensor can be considered as three independent observations. The posterior of the previous tapping and the X-axis LR data are used to update the posterior of the X-axis. Then, this posterior serves as the prior for the Y-axis, resulting in a sequential state update.}
	\label{seq_filter}
        \vspace{-1em} 
 
\end{figure}

Initializing the state probability distribution with the Gaussian prior, we have
\begin{equation}
    \begin{split}
    \mu_0 = 0, \
    \mathbf{\Sigma}_0(i,j)  = A \ \mathrm{exp}\left\{  - \frac{||\mathbf{v}_i - \mathbf{v}_j||^2}{r^2}  \right\}.
    \end{split}
\end{equation}

Given all the observations $\mathbf{I}^{\mathrm{LR}}_{1:t}$, the clip matrix $\mathbf{C}_{1:t}$ and the degradation $\mathbf{H}(\gamma)$, we can obtain based on the Bayesian formula that

\begin{equation}
    \begin{split}
    P(\mathbf{S}_t | \mathbf{I}^{\mathrm{LR}}_{1:t}, \mathbf{C}_{1:t}, \mathbf{H}(\gamma)) = \eta  & P(\mathbf{I}^{\mathrm{LR}}_t | \mathbf{S}_t, \mathbf{I}^{\mathrm{LR}}_{1:t-1}, \mathbf{C}_{1:t}, \mathbf{H}(\gamma)) \cdot \\
    & P(\mathbf{S}_{t} | \mathbf{I}^{\mathrm{LR}}_{1:t-1}, \mathbf{C}_{1:t}, \mathbf{H}(\gamma))
    \label{eq:state_bayesain_1}
    \end{split}.
\end{equation}

The degradation matrix $\mathbf{H}(\gamma)$ represents the intrinsic property of the sensor, independent of state and observation variables. Moreover, $\mathbf{I}^{\mathrm{LR}}_t \perp \!\!\! \perp \mathbf{I}^{\mathrm{LR}}_{1:t-1}$, $\mathbf{I}^{\mathrm{LR}}_t \perp \!\!\! \perp \mathbf{C}_{1:t-1}$, $\mathbf{S}_t \perp \!\!\! \perp \mathbf{C}_t$. In addition, the posterior probability of $\mathbf{S}_{t-1}$ can directly serve as the prior probability of $\mathbf{S}_{t}$ as indicated in Eq.\ref{eq:state_func}. Thus, Eq.\ref{eq:state_bayesain_1} can be simplified as 

\begin{equation}
    \begin{split}
    P(\mathbf{S}_t | \mathbf{I}^{\mathrm{LR}}_{1:t}, \mathbf{C}_{1:t}) = \eta P(\mathbf{I}^{\mathrm{LR}}_t | \mathbf{S}_t, \mathbf{C}_{t}) P(\mathbf{S}_{t-1}|\mathbf{I}^{\mathrm{LR}}_{1:t-1}, \mathbf{C}_{1:t-1}) 
    \end{split},
    \label{eq:state_bayesain_2}
\end{equation}
where $ P(\mathbf{S}_{t-1}|\mathbf{I}^{\mathrm{LR}}_{1:t-1}, \mathbf{C}_{1:t-1}) $ is the posterior probability of the ($t-1$)-th tapping, encompassing all information prior to the $t$-th tapping. Its probability distribution is

\begin{equation}
    P(\mathbf{S}_{t-1}|\mathbf{I}^{\mathrm{LR}}_{1:t-1}, \mathbf{C}_{1:t-1})  \sim \mathcal{N}(\mu_{t-1}, \mathbf{\Sigma}_{t-1}).
\end{equation}

The $P(\mathbf{I}^{\mathrm{LR}}_t | \mathbf{S}_t, \mathbf{C}_{t})$ in Eq. \ref{eq:state_bayesain_2} is the likelihood probability, which is determined by the degradation process (sensor model) as described in Eq. (\ref{eq:clip_matrix}, \ref{eq:degradation_matrix}, \ref{eq:xy_gradient}). In this paper,the sensor collects   3-axis LR tactile data at each instance as $\mathbf{I}^{\mathrm{LR}}_t = [\mathbf{I}^{\mathrm{LR}}_{x,t}, \mathbf{I}^{\mathrm{LR}}_{y,t}, \mathbf{I}^{\mathrm{LR}}_{z,t}]$. We assume that the collected 3-axis data are mutually independent, and can be regarded as the sampling results obtained by three independent sensors simultaneously sampling the contact surface. For this type of synchronously collected data in a multi-sensor Kalman filter, \cite{multi-sensor-KF} provides a systematic description. We employ a sequential filter to the update state, which means that the observation data will be processed sequentially, as shown in  Fig.\ref{seq_filter}. Here, we only demonstrate the update formulas for the X-axis, and those of Y- and Z-axis can be obtained similarly.

\begin{equation}
    \begin{split}
    \mathbf{\Sigma}_{x|t} &= [(\mathbf{H}_x\mathbf{G}_x\mathbf{C}_t)^T \mathbf{Q}^{-1}_{x, t}\mathbf{H}_x\mathbf{G}_x\mathbf{C}_t + \mathbf{\Sigma}^{-1}_{t-1}]^{-1} \\
    \mathbf{K}_{x|t} &= \mathbf{\Sigma}_{x|t} (\mathbf{H}_x\mathbf{G}_x)^T \mathbf{Q}^{-1}_{x, t} \\
    \mathbf{\mu}_{x|t} &= \mathbf{\mu}_{t-1} + \mathbf{K}_{x|t}(\mathbf{I}^{\mathrm{LR}}_{x, t} - \mathbf{H}_x\mathbf{G}_x\mathbf{C}_t\mu_{t-1})
    \end{split},
\end{equation}
where $\mathbf{K}_{x|t}$ is the Kalman gain of X-axis data. This way, we can reconstruct the tactile pattern iteratively with the above developed Kalman filter.

\section{Active Tactile Exploration Policy}
In this section, we present a heuristic active tactile exploration strategy to improve the efficiency of the proposed reconstruction approach. The developed strategy is motivated by the human tactile exploration process as mentioned in Section \ref{section:intro}, where both the contour information and the uncertainty of the currently reconstructed contact surface are considered for effective tactile exploration. In this research, the state $\mathbf{S}_t$, describes the shape of the contact surface, which can be divided into four categories. (A) is probably not a contour; (B) is probably a contour; (C) may not be a contour; (D) may be a contour. In the next action $\mathbf{m}_{t+1}$, we should focus on primarily on region of type (D), as the information in this region will directly determine the quality of the reconstruction. Naturally, we should reduce the ineffective sampling in the region of type (A). This way, we can develop a tactile exploration algorithm to effectively enhance the reconstruction quality.

We calculate the gradient of the current state, $ \nabla \mu_t $, which reflects the contour information. The larger the gradient, the more prominent the contour information. Consequently, we define a \textit{gradient map} $\mathbf{G}_t \in \mathbb{R}^{(\alpha \cdot M)^2\times 1}$

\begin{equation}
    \begin{split}
        \mathbf{G}_t = (1-e^{-\lambda t})\nabla\mu_t + e^{-\lambda t}
    \end{split},
    \label{eq:gradient_map}
\end{equation}
where $\lambda$ is a hyper-parameter. Additionally, we define a \textit{uncertain map} $\mathbf{U}_t \in \mathbb{R}^{(\alpha \cdot M)^2\times 1}$ to describe the uncertainty of the state, where

\begin{equation}
    \begin{split}
        \mathbf{U}_t = \frac{1}{2} \log (2\pi \ \mathrm{diag}(\mathbf{\Sigma}_t)) + \frac{1}{2}
    \end{split}.
\end{equation}

The variance of each variable in the covariance matrix $\mathbf{\Sigma}_t$ is chosen to describe the uncertainty. Subsequently, we define the \textit{decision map} for decision-making.

\begin{equation}
    \begin{split}
        \mathbf{D}_{t} = \mathbf{G}_t  \odot \mathbf{U}_t
    \end{split},
\end{equation}
where $\odot$ denotes the corresponding elemental product. The proposed heuristic exploration strategy is an \textit{explore-then-exploit} strategy. At $t=0$, the sensor conducts its initial tapping, with $\mathbf{D}_t = \mathbf{U}_t$, marking as the explore phase. As $t$ increases, $\mathbf{D}_t \rightarrow  \nabla \mu_t  \mathbf{U}_t$. During decision-making, we take both the current contour information and uncertainty rate into account, denoted as the exploit phase. The hyper-parameter $\lambda$ in Eq.\ref{eq:gradient_map} can be considered as the transition rate from the explore phase to the exploit phase, which is set to 0.7 in this paper. The selection of $\lambda$ is task related, and can be obtained generally through experiments. The next action $\mathbf{m}_{t+1}$ should satisfy
\begin{equation}
    \begin{split}
    \mathbf{\hat{m}}_{t+1} =  \arg \underset{\mathbf{m}_{t+1} \in \mathcal{M}}{\max}  \sum_0^{M^2} \mathbf{C}_t \mathbf{D}_t   
    \end{split},
\end{equation}
which indicates that the next movement should be the region with the largest sum value enclosing $\mathbf{D}_t $. As a result, we can manage an effective tactile exploration along with the reconstruction process, as shown in Fig.\ref{tap_system}.

\section{Experimental Results}

\begin{figure}[t]
	\centering
	\includegraphics[width=0.43\textwidth]{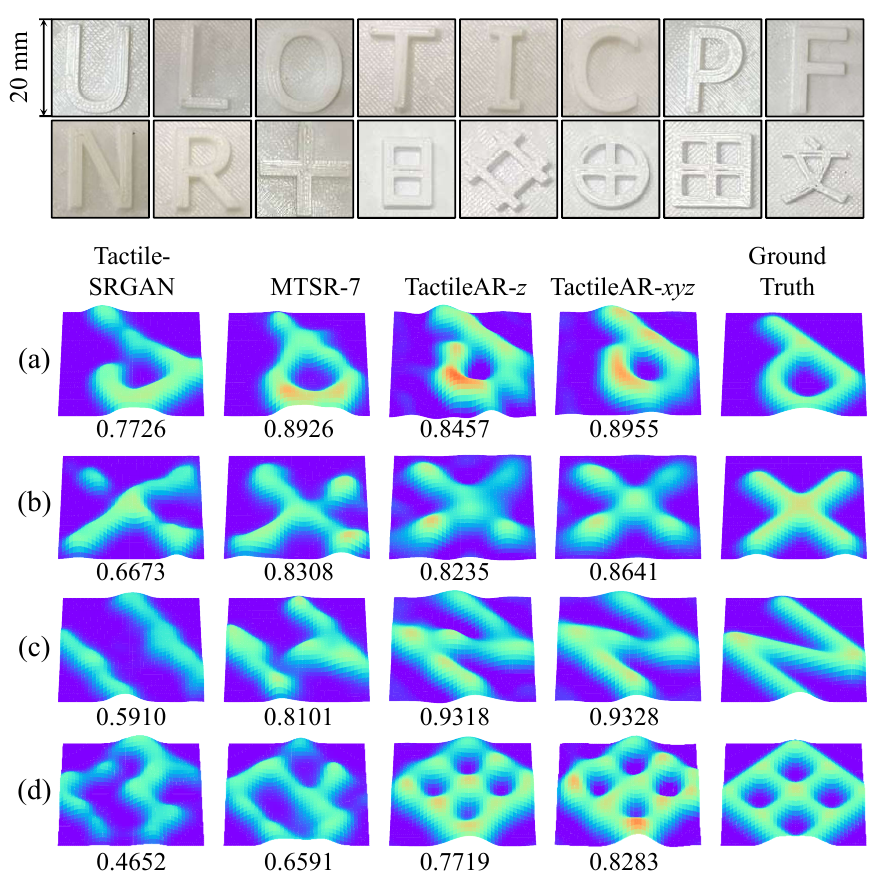}
	\caption{Experimental results of predicted HR data. The 8 contact surfaces in the bottom row are not included in the training dataset of the tactile SR model (TactileSRGAN\cite{wubing_tactile_sr_iros_2022}, MTSR-7\cite{wubing_tactile_sr_toh_2023}), while the 8 above are included, but their poses are not encompassed in the training dataset. TactileAR-$z$ and TactileAR-$xyz$ denote the results after tapping the surface for 20 times using LR data in the Z-axis and XYZ-axis, respectively, with the active exploration strategy ($\lambda=0.7$). Below the pictures are the corresponding SSIM results.}
	\label{fig:SR_result_vis}
        \vspace{-1.5em} 
 
\end{figure}

\subsection{System Setups}
In this paper, we employ the Xela tactile sensor\cite{xela-tactile-sensor}, a commercial 3-axis Hall tactile sensor as shown in Fig.\ref{fig:xela_tactile_sensor}. The sensor has $4\times 4$ elements (taxel) with \texttildemid 20 $\times$ 20 mm$^2$ sensing areas and is mounted on the end effector of an Aubo-i5 robot arm from AUBO robotics.  Each taxel can collect 3D deformation information of a specific region decoupled into X-, Y- and Z-axis. We utilize the raw sensor output data and scale the value to $[0,1]$ without further calibration.  The sensor data and the applied force correspond to each other and more precise mapping is delineated in Figs. 7-8 in \cite{xela-tactile-sensor}.

During the $t$-th tapping procedure, the sensor taps the contact surface vertically downward and reaches a given height to collect a LR  tactile pattern $\mathbf{I}^{\mathrm{LR}}_t$and corresponding position $\mathbf{m}_t$. Then the sensor returns to the initial position, waiting for the next tapping. In the experiment, we discrete the motion of the sensor, with the center of the sensor consistently moving within the reconstructed region. $\mathbf{m}_t \in \mathcal{M}=\mathcal{X} \times \mathcal{Y}\times \Theta$ where $ \mathcal{X}, \mathcal{Y} = \{-\frac{l^{state}}{2}, -\frac{l^{state}}{2}+dx, \cdots, \frac{l^{state}}{2} \}, \Theta = \{-\frac{\pi}{2},  -\frac{\pi}{2}+d\theta, \cdots, \frac{\pi}{2}\}$. The translation step size is $dx=0.5mm$ and the rotational step size is $d\theta=5^{\circ}$, which are much larger than the motion accuracy ($\pm 0.03mm$) of the robot arm. Therefore, we ignore the effect of motion errors of the robot arm.

\begin{figure*}[h]
	\centering
	\includegraphics[width=0.80\textwidth]{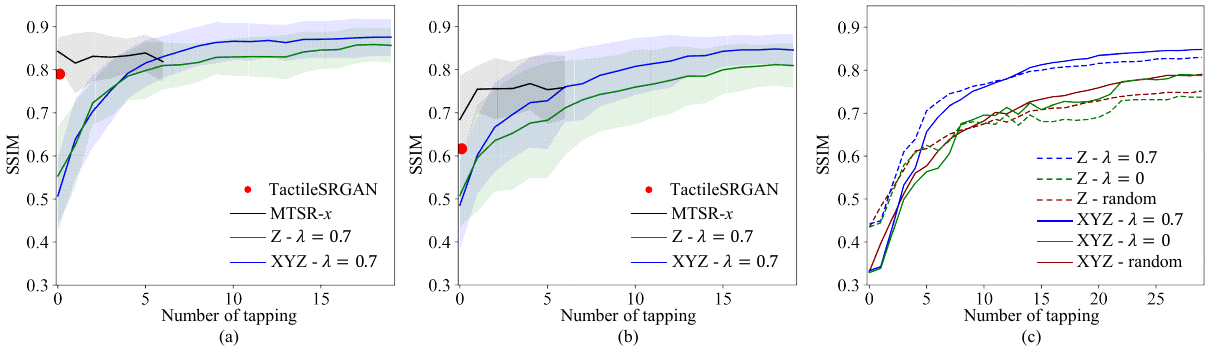}
	\caption{Reconstruction performance of the TactileAR. (a) and (b) illustrate the comparison results when contact surfaces are included or excluded in the SR model training set, respectively. (c) indicates the performance of TactileAR with different tapping policies and input data. MTSR-x denotes collecting LR data along a fixed trajectory for x (  \textless 7) times, then predicts the HR data for the last tapping.  Z-$\lambda=0.7$ indicates that TactileAR uses Z-axis observations and the proposed active tapping policy with $\lambda=0.7$.}
	\label{fig:result_curve_3}
        \vspace{-1.5em} 
 
\end{figure*}

\begin{figure}[h]
	\centering
	\includegraphics[width=0.43\textwidth]{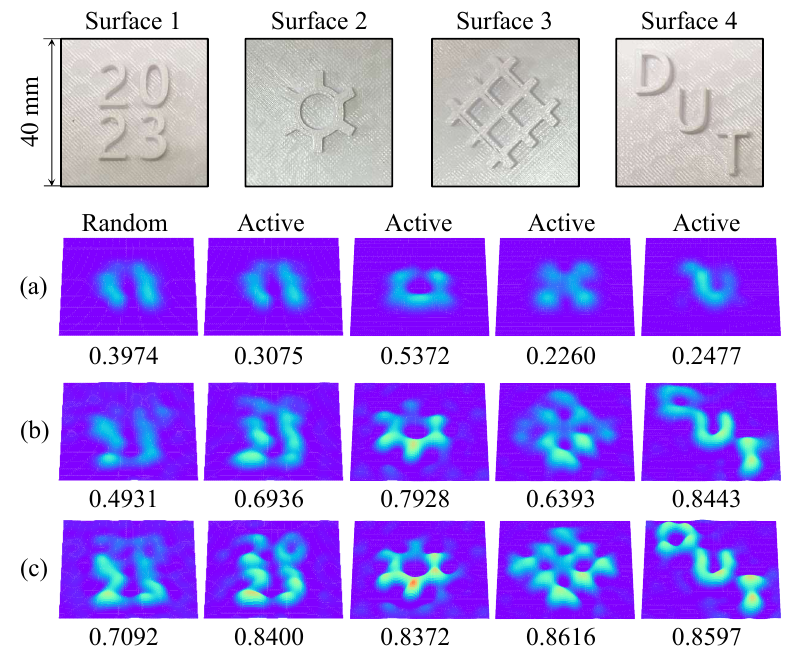}
	\caption{Experimental results for complex contact surface. `random` and `active` denote the sampling policy.  (a), (b) and (c) represent the reconstruction results of triaxial data for 1, 10 and 30 tappings to the contact surface, respectively. The corresponding SSIM results are displayed below the figures.}
	\label{fig:active_recon_vis}
        \vspace{-1.5em} 
 
\end{figure}

\subsection{HR Tactile Data Prediction}
The TactileAR is employed to predict HR data of contact surfaces. Based on Eq.\ref{eq:clip_matrix}, we can estimate the corresponding HR data $\mathbf{I}^{\mathrm{HR}}_t$ of the sensor based on the current estimated state $\mathbf{S}_t$ and position $\mathbf{m}_t$.   We compare the performance of TactileAR, which only relies on the observation data, with existing methods relying on the prior information \cite{wubing_tactile_sr_iros_2022, wubing_tactile_sr_toh_2023} e.g. learning-based algorithms which highly rely on pre-design datasets. In order to present a fair comparison, we adopt 16 contact surfaces, of which  8 contact surfaces are included in the dataset, but their poses are not encompassed, while 8 are excluded from the dataset, as depicted in Fig.\ref{fig:SR_result_vis}. The HR data predicted by \cite{wubing_tactile_sr_iros_2022, wubing_tactile_sr_toh_2023} contains force and shape information, and we scale the data to $[0, 1]$ so that it can be compared with the proposed algorithm.

The MTSR-$x$ is proposed in \cite{wubing_tactile_sr_toh_2023} which is a SOTA tactile SR model for predicting HR data of contact surfaces based on prior knowledge. The MTSR-$x$ collects LR data of contact surface along a fixed trajectory for $x$ times and then predicts the HR data for the last tapping. It is trained using a tactile SR dataset. The black curves in Fig. \ref{fig:result_curve_3}(a,b) show the performance of the SR model for contact surfaces included and excluded in the training dataset, respectively. We can observe that the SR models which require the prior information  only need a small amount of observations to recover high-quality shapes. However, their performance is limited by the dataset used for model training, which decreases for the contact surfaces not in the dataset.

The TactileAR only uses the observation data  to reconstruct the HR data of the contact surface, which implies that the TactileAR does not rely on the prior information. Hence, the TactileAR will not suffer from the generalization problem shown in learning-based methods. We employ the proposed exploration policy to tapping the contact surface for 20 times and predict the corresponding HR data after each tapping. The blue and red curves in Fig.\ref{fig:result_curve_3}(a, b) show the performance. We also evaluate the impact of using Z-axis versus XYZ-axis observation data. With the increasing number of tappings, the reconstruction performance of using three-axis data becomes slightly superior to that of the single-axis case.

\subsection{Active Reconstruction Performance}
We also evaluate the performance of the TacatileAR in the reconstruction of complex contact surface shapes. We set $\alpha=2, N=4, M=40$. This indicates that the TactileAR reconstructs the shape of contact surface with 4-times bigger than the sensor  (i.e. $40\times40$ mm$^2$), with a resolution of $80 \times 80$. Experimental results are illustrated in Fig.\ref{fig:active_recon_vis}. Here, three exploration policies are evaluated. `random' means the sensor randomly choose a tapping position from motion space $\mathcal{M}$. `$\lambda=0$' indicates that the sensor only considers the uncertainty of the system during sampling process, which can be considered as the pure active exploration policy, its sampling trajectory is fixed in this paper, and does not depend on the shape of the contact surface. In contrast, the proposed active exploration policy combines the contour with the uncertainty, and therefore improves the reconstruction efficiency, as shown in Fig.\ref{fig:result_curve_3}(c) and Fig.\ref{fig:active_recon_vis}.


\section{Conclusion}

In this paper, we focus on reconstructing HR contact surface shapes using LR tactile sensor through a framework combing the Gaussian degradation model and the Kalman filter, alongside a heuristic active exploration strategy that considers contour and surface uncertainty. Our algorithm, evaluated against prior-information-based methods in real-world cases, demonstrates superior ability to reconstruct high-quality contact surfaces from minimal LR observation data. Furthermore, it exhibits exceptional performance in recovering complex contact surfaces compared with the prior-information-based algorithms.

The limitation of our method is its reliance on the premise that objects do not deform or move during tapping, which narrows its field of application. In the future, we plan to refine our algorithm to lessen dependence on this assumption and improve reconstruction efficiency by substituting \textit{tapping} with \textit{sliding}.


\nocite{*}


\bibliographystyle{IEEEtran}
\balance
\bibliography{reference.bib}

\begin{thebibliography}{10}
\providecommand{\url}[1]{#1}
\csname url@samestyle\endcsname
\providecommand{\newblock}{\relax}
\providecommand{\bibinfo}[2]{#2}
\providecommand{\BIBentrySTDinterwordspacing}{\spaceskip=0pt\relax}
\providecommand{\BIBentryALTinterwordstretchfactor}{4}
\providecommand{\BIBentryALTinterwordspacing}{\spaceskip=\fontdimen2\font plus
\BIBentryALTinterwordstretchfactor\fontdimen3\font minus \fontdimen4\font\relax}
\providecommand{\BIBforeignlanguage}[2]{{%
\expandafter\ifx\csname l@#1\endcsname\relax
\typeout{** WARNING: IEEEtran.bst: No hyphenation pattern has been}%
\typeout{** loaded for the language `#1'. Using the pattern for}%
\typeout{** the default language instead.}%
\else
\language=\csname l@#1\endcsname
\fi
#2}}
\providecommand{\BIBdecl}{\relax}
\BIBdecl

\bibitem{human_tactile_sr}
J.~M. Loomis, ``An investigation of tactile hyperacuity,'' \emph{Sensory processes}, vol.~3, no.~4, pp. 289--302, 1979.

\bibitem{tactile_sr_2015}
N.~F. Lepora, U.~Martinez-Hernandez, M.~Evans, L.~Natale, G.~Metta, and T.~J. Prescott, ``Tactile superresolution and biomimetic hyperacuity,'' \emph{IEEE Transactions on Robotics}, vol.~31, no.~3, pp. 605--618, 2015.

\bibitem{gelsight_2017_sensor}
W.~Yuan, S.~Dong, and E.~H. Adelson, ``Gelsight: High-resolution robot tactile sensors for estimating geometry and force,'' \emph{Sensors}, vol.~17, no.~12, p. 2762, 2017.

\bibitem{gelslim_2018_IROS}
E.~Donlon, S.~Dong, M.~Liu, J.~Li, E.~Adelson, and A.~Rodriguez, ``Gelslim: A high-resolution, compact, robust, and calibrated tactile-sensing finger,'' in \emph{2018 IEEE/RSJ International Conference on Intelligent Robots and Systems (IROS)}.\hskip 1em plus 0.5em minus 0.4em\relax IEEE, 2018, pp. 1927--1934.

\bibitem{tactip_2018_SoftRobot}
B.~Ward-Cherrier, N.~Pestell, L.~Cramphorn, B.~Winstone, M.~E. Giannaccini, J.~Rossiter, and N.~F. Lepora, ``The tactip family: Soft optical tactile sensors with 3d-printed biomimetic morphologies,'' \emph{Soft robotics}, vol.~5, no.~2, pp. 216--227, 2018.

\bibitem{vision-sensor-2022}
H.~Sun, K.~J. Kuchenbecker, and G.~Martius, ``A soft thumb-sized vision-based sensor with accurate all-round force perception,'' \emph{Nature Machine Intelligence}, vol.~4, no.~2, pp. 135--145, 2022.

\bibitem{tactip-review-2020}
N.~F. Lepora, ``Soft biomimetic optical tactile sensing with the tactip: A review,'' \emph{IEEE Sensors Journal}, 2021.

\bibitem{soft_vit_2022_RoboSoft}
S.~Q. Liu and E.~H. Adelson, ``Gelsight fin ray: Incorporating tactile sensing into a soft compliant robotic gripper,'' in \emph{2022 IEEE 5th International Conference on Soft Robotics (RoboSoft)}.\hskip 1em plus 0.5em minus 0.4em\relax IEEE, 2022, pp. 925--931.

\bibitem{mems-sensor-2008}
H.-K. Lee, J.~Chung, S.-I. Chang, and E.~Yoon, ``Normal and shear force measurement using a flexible polymer tactile sensor with embedded multiple capacitors,'' \emph{Journal of Microelectromechanical Systems}, vol.~17, no.~4, pp. 934--942, 2008.

\bibitem{mems-sensor-2015}
A.~Charalambides and S.~Bergbreiter, ``A novel all-elastomer mems tactile sensor for high dynamic range shear and normal force sensing,'' \emph{Journal of Micromechanics and Microengineering}, vol.~25, no.~9, p. 095009, 2015.

\bibitem{tactile_sensor_review_2015_RAS}
Z.~Kappassov, J.-A. Corrales, and V.~Perdereau, ``Tactile sensing in dexterous robot hands,'' \emph{Robotics and Autonomous Systems}, vol.~74, pp. 195--220, 2015.

\bibitem{tactile_sr_2022_ral}
Y.~Yan, Y.~Shen, C.~Song, and J.~Pan, ``Tactile super-resolution model for soft magnetic skin,'' \emph{IEEE Robotics and Automation Letters}, vol.~7, no.~2, pp. 2589--2596, 2022.

\bibitem{tactile_sr_2015_iros}
N.~F. Lepora and B.~Ward-Cherrier, ``Superresolution with an optical tactile sensor,'' in \emph{2015 IEEE/RSJ International Conference on Intelligent Robots and Systems (IROS)}.\hskip 1em plus 0.5em minus 0.4em\relax IEEE, 2015, pp. 2686--2691.

\bibitem{tactile_sr_2022_sr}
H.~Sun and G.~Martius, ``Guiding the design of superresolution tactile skins with taxel value isolines theory,'' \emph{Science Robotics}, vol.~7, no.~63, p. eabm0608, 2022.

\bibitem{wubing_tactile_sr_iros_2022}
B.~Wu, Q.~Liu, and Q.~Zhang, ``Tactile pattern super resolution with taxel-based sensors,'' in \emph{2022 IEEE/RSJ International Conference on Intelligent Robots and Systems (IROS)}.\hskip 1em plus 0.5em minus 0.4em\relax IEEE, 2022, pp. 3644--3650.

\bibitem{wubing_tactile_sr_toh_2023}
B.~Wu and Q.~Liu, ``Integrating point spread function into taxel-based tactile pattern super resolution.'' \emph{IEEE Transactions on Haptics}, 2024.

\bibitem{tactile_pattern_review_2017}
S.~Luo, J.~Bimbo, R.~Dahiya, and H.~Liu, ``Robotic tactile perception of object properties: A review,'' \emph{Mechatronics}, vol.~48, pp. 54--67, 2017.

\bibitem{tactile_perception_review_2020_TRO}
Q.~Li, O.~Kroemer, Z.~Su, F.~F. Veiga, M.~Kaboli, and H.~J. Ritter, ``A review of tactile information: Perception and action through touch,'' \emph{IEEE Transactions on Robotics}, vol.~36, no.~6, pp. 1619--1634, 2020.

\bibitem{tactile_active_2013_rss}
N.~F. Lepora, U.~Martinez-Hernandez, and T.~J. Prescott, ``Active bayesian perception for simultaneous object localization and identification.'' in \emph{Robotics: Science and Systems}.\hskip 1em plus 0.5em minus 0.4em\relax Citeseer, 2013, pp. 1--8.

\bibitem{tactile_mapping_2011}
Z.~Pezzementi, C.~Reyda, and G.~D. Hager, ``Object mapping, recognition, and localization from tactile geometry,'' in \emph{2011 IEEE International Conference on Robotics and Automation}.\hskip 1em plus 0.5em minus 0.4em\relax IEEE, 2011, pp. 5942--5948.

\bibitem{active_pure_2016_iros}
Z.~Yi, R.~Calandra, F.~Veiga, H.~van Hoof, T.~Hermans, Y.~Zhang, and J.~Peters, ``Active tactile object exploration with gaussian processes,'' in \emph{2016 IEEE/RSJ International Conference on Intelligent Robots and Systems (IROS)}.\hskip 1em plus 0.5em minus 0.4em\relax IEEE, 2016, pp. 4925--4930.

\bibitem{active_pure_2013_ICRA}
D.~Xu, G.~E. Loeb, and J.~A. Fishel, ``Tactile identification of objects using bayesian exploration,'' in \emph{2013 IEEE international conference on robotics and automation}.\hskip 1em plus 0.5em minus 0.4em\relax IEEE, 2013, pp. 3056--3061.

\bibitem{active_heur_tactile_2022_tro}
C.~Xiao, S.~Xu, W.~Wu, and J.~Wachs, ``Active multiobject exploration and recognition via tactile whiskers,'' \emph{IEEE Transactions on Robotics}, vol.~38, no.~6, pp. 3479--3497, 2022.

\bibitem{active_heur_tactile_2011_tro}
Z.~Pezzementi, E.~Plaku, C.~Reyda, and G.~D. Hager, ``Tactile-object recognition from appearance information,'' \emph{IEEE Transactions on robotics}, vol.~27, no.~3, pp. 473--487, 2011.

\bibitem{active_heur_tactile_2017_ral}
N.~F. Lepora, K.~Aquilina, and L.~Cramphorn, ``Exploratory tactile servoing with active touch,'' \emph{IEEE Robotics and Automation Letters}, vol.~2, no.~2, pp. 1156--1163, 2017.

\bibitem{active_rl_tactile_2020_ral}
A.~Church, J.~Lloyd, R.~Hadsell, and N.~F. Lepora, ``Deep reinforcement learning for tactile robotics: Learning to type on a braille keyboard,'' \emph{IEEE Robotics and Automation Letters}, vol.~5, no.~4, pp. 6145--6152, 2020.

\bibitem{active_rl_tactile_2022_iros}
S.~Jiang and L.~L. Wong, ``Active tactile exploration using shape-dependent reinforcement learning,'' in \emph{2022 IEEE/RSJ International Conference on Intelligent Robots and Systems (IROS)}.\hskip 1em plus 0.5em minus 0.4em\relax IEEE, 2022, pp. 8995--9002.

\bibitem{tactile_sim_to_real_review}
Y.~Narang, B.~Sundaralingam, M.~Macklin, A.~Mousavian, and D.~Fox, ``Sim-to-real for robotic tactile sensing via physics-based simulation and learned latent projections,'' in \emph{2021 IEEE International Conference on Robotics and Automation (ICRA)}.\hskip 1em plus 0.5em minus 0.4em\relax IEEE, 2021, pp. 6444--6451.

\bibitem{vision_tactile_map_2019_ICRA}
M.~Bauza, O.~Canal, and A.~Rodriguez, ``Tactile mapping and localization from high-resolution tactile imprints,'' in \emph{2019 International Conference on Robotics and Automation (ICRA)}.\hskip 1em plus 0.5em minus 0.4em\relax IEEE, 2019, pp. 3811--3817.

\bibitem{DenseTact_2022_ICRA}
W.~K. Do and M.~Kennedy, ``Densetact: Optical tactile sensor for dense shape reconstruction,'' in \emph{2022 International Conference on Robotics and Automation (ICRA)}.\hskip 1em plus 0.5em minus 0.4em\relax IEEE, 2022, pp. 6188--6194.

\bibitem{vision_tactile_reconstruction_2018_IROS}
S.~Wang, J.~Wu, X.~Sun, W.~Yuan, W.~T. Freeman, J.~B. Tenenbaum, and E.~H. Adelson, ``3d shape perception from monocular vision, touch, and shape priors,'' in \emph{2018 IEEE/RSJ International Conference on Intelligent Robots and Systems (IROS)}.\hskip 1em plus 0.5em minus 0.4em\relax IEEE, 2018, pp. 1606--1613.

\bibitem{visual_taxle_reconstruction_2013_IROS}
M.~Bj{\"o}rkman, Y.~Bekiroglu, V.~H{\"o}gman, and D.~Kragic, ``Enhancing visual perception of shape through tactile glances,'' in \emph{2013 IEEE/RSJ International Conference on Intelligent Robots and Systems}.\hskip 1em plus 0.5em minus 0.4em\relax IEEE, 2013, pp. 3180--3186.

\bibitem{visual_taxle_reconstruction_2014_IJRR}
J.~Ilonen, J.~Bohg, and V.~Kyrki, ``Three-dimensional object reconstruction of symmetric objects by fusing visual and tactile sensing,'' \emph{The International Journal of Robotics Research}, vol.~33, no.~2, pp. 321--341, 2014.

\bibitem{tactile_reconstruction_2011_TRO_short}
M.~Meier, M.~Schopfer, R.~Haschke, and H.~Ritter, ``A probabilistic approach to tactile shape reconstruction,'' \emph{IEEE Transactions on Robotics}, vol.~27, no.~3, pp. 630--635, 2011.

\bibitem{GPIS_2017_IROS}
D.~Driess, P.~Englert, and M.~Toussaint, ``Active learning with query paths for tactile object shape exploration,'' in \emph{2017 IEEE/RSJ international conference on intelligent robots and systems (IROS)}.\hskip 1em plus 0.5em minus 0.4em\relax IEEE, 2017, pp. 65--72.

\bibitem{GPIS_2019_ICRA}
D.~Driess, D.~Hennes, and M.~Toussaint, ``Active multi-contact continuous tactile exploration with gaussian process differential entropy,'' in \emph{2019 International Conference on Robotics and Automation (ICRA)}.\hskip 1em plus 0.5em minus 0.4em\relax IEEE, 2019, pp. 7844--7850.

\bibitem{cv-gauss-degradation-cvpr-2018}
K.~Zhang, W.~Zuo, and L.~Zhang, ``Learning a single convolutional super-resolution network for multiple degradations,'' in \emph{Proceedings of the IEEE conference on computer vision and pattern recognition}, 2018, pp. 3262--3271.

\bibitem{multi-sensor-KF}
D.~Willner, C.-B. Chang, and K.-P. Dunn, ``Kalman filter algorithms for a multi-sensor system,'' in \emph{1976 IEEE conference on decision and control including the 15th symposium on adaptive processes}.\hskip 1em plus 0.5em minus 0.4em\relax IEEE, 1976, pp. 570--574.

\bibitem{xela-tactile-sensor}
T.~P. Tomo, S.~Somlor, A.~Schmitz, L.~Jamone, W.~Huang, H.~Kristanto, and S.~Sugano, ``Design and characterization of a three-axis hall effect-based soft skin sensor,'' \emph{Sensors}, vol.~16, no.~4, p. 491, 2016.

\bibitem{EIT_sensor_2021_TRO}
H.~Park, K.~Park, S.~Mo, and J.~Kim, ``Deep neural network based electrical impedance tomographic sensing methodology for large-area robotic tactile sensing,'' \emph{IEEE Transactions on Robotics}, vol.~37, no.~5, pp. 1570--1583, 2021.

\bibitem{tactile_sensor_review_2017_sensor}
L.~Zou, C.~Ge, Z.~J. Wang, E.~Cretu, and X.~Li, ``Novel tactile sensor technology and smart tactile sensing systems: A review,'' \emph{Sensors}, vol.~17, no.~11, p. 2653, 2017.

\bibitem{Biotac_2008}
J.~A. Fishel, V.~J. Santos, and G.~E. Loeb, ``A robust micro-vibration sensor for biomimetic fingertips,'' in \emph{2008 2nd IEEE RAS \& EMBS International Conference on Biomedical Robotics and Biomechatronics}.\hskip 1em plus 0.5em minus 0.4em\relax IEEE, 2008, pp. 659--663.

\bibitem{tactile-review-2009-tro}
R.~S. Dahiya, G.~Metta, M.~Valle, and G.~Sandini, ``Tactile sensing—from humans to humanoids,'' \emph{IEEE transactions on robotics}, vol.~26, no.~1, pp. 1--20, 2009.

\bibitem{li2020review}
Q.~Li, O.~Kroemer, Z.~Su, F.~F. Veiga, M.~Kaboli, and H.~J. Ritter, ``A review of tactile information: Perception and action through touch,'' \emph{IEEE Transactions on Robotics}, vol.~36, no.~6, pp. 1619--1634, 2020.

\end{thebibliography}

\end{document}